\documentclass[sigconf]{acmart}

\usepackage{graphicx}
\usepackage{amsmath}
\usepackage{booktabs}
\usepackage{multirow}
\usepackage{color}
\usepackage{multicol}

\AtBeginDocument{%
  \providecommand\BibTeX{{%
    \normalfont B\kern-0.5em{\scshape i\kern-0.25em b}\kern-0.8em\TeX}}}

\renewcommand\footnotetextcopyrightpermission[1]{}
\setcopyright{acmcopyright}
\copyrightyear{2023}
\acmYear{2023}
\acmDOI{XXXXXXX.XXXXXXX}

\acmBooktitle{Preprint Version} 

\acmSubmissionID{251}

\begin{document}

\title{Hierarchical Dynamic Image Harmonization}


\author{Haoxing Chen}
\affiliation{%
 \institution{Nanjing University}
 \institution{Tiansuan Lab, Ant Group Inc.}
 \city{Hangzhou}
 \country{China}}
 \email{hx.chen@hotmail.com}

\author{Zhangxuan Gu}
\affiliation{%
 \institution{Tiansuan Lab, Ant Group Inc.}
 \city{Shanghai}
 \country{China}}
 \email{guzhangxuan.gzx@antgroup.com}

\author{Yaohui Li}
\affiliation{%
 \institution{Nanjing University}
 \city{Nanjing}
 \country{China}}
 \email{yaohuili@smail.nju.edu.cn}

\author{Jun Lan}
\affiliation{%
 \institution{Tiansuan Lab, Ant Group Inc.}
 \city{Hangzhou}
 \country{China}}
 \email{yelan.lj@antgroup.com}

 \author{Changhua Meng}
\affiliation{%
 \institution{Tiansuan Lab, Ant Group Inc.}
 \city{Hangzhou}
 \country{China}}
 \email{changhua.mch@antgroup.com}

 \author{Weiqiang Wang}
\affiliation{%
 \institution{Tiansuan Lab, Ant Group Inc.}
 \city{Hangzhou}
 \country{China}}
 \email{weiqiang.wwq@antgroup.com}

 \author{Huaxiong Li}
\affiliation{%
 \institution{Department of Control Science and Intelligence Engineering, Nanjing University}
 \city{Nanjing}
 \country{China}}
 \email{huaxiongli@nju.edu.cn}


\renewcommand{\shortauthors}{Chen et al.}

\begin{abstract}
Image harmonization is a critical task in computer vision, which aims to adjust the foreground to make it compatible with the background. Recent works mainly focus on using global transformations (i.e., normalization and color curve rendering) to achieve visual consistency. However, these models ignore local visual consistency and their huge model sizes limit their harmonization ability on edge devices. In this paper, we propose a hierarchical dynamic network (HDNet) to adapt features from local to global view for better feature transformation in efficient image harmonization. Inspired by the success of various dynamic models, local dynamic (LD) module and mask-aware global dynamic (MGD) module are proposed in this paper. Specifically, LD matches local representations between the foreground and background regions based on semantic similarities, then adaptively adjust every foreground local representation according to the appearance of its $K$-nearest neighbor background regions. In this way, LD can produce more realistic images at a more fine-grained level, and simultaneously enjoy the characteristic of semantic alignment. The MGD effectively applies distinct convolution to the foreground and background, learning the representations of foreground and background regions as well as their correlations to the global harmonization, facilitating local visual consistency for the images much more efficiently. Experimental results demonstrate that the proposed HDNet significantly reduces the total model parameters by more than 80\% compared to previous methods, while still attaining state-of-the-art performance on the popular iHarmony4 dataset. Additionally, we introduced a lightweight version of HDNet, i.e., HDNet-lite, which has only 0.65MB parameters, yet it still achieved competitive performance. Our code is avaliable in \url{https://github.com/chenhaoxing/HDNet}.
\end{abstract}





\maketitle

\section{Introduction}

\begin{figure}[t]
	\centering
	\includegraphics[height=11.3cm,width=8.6cm]{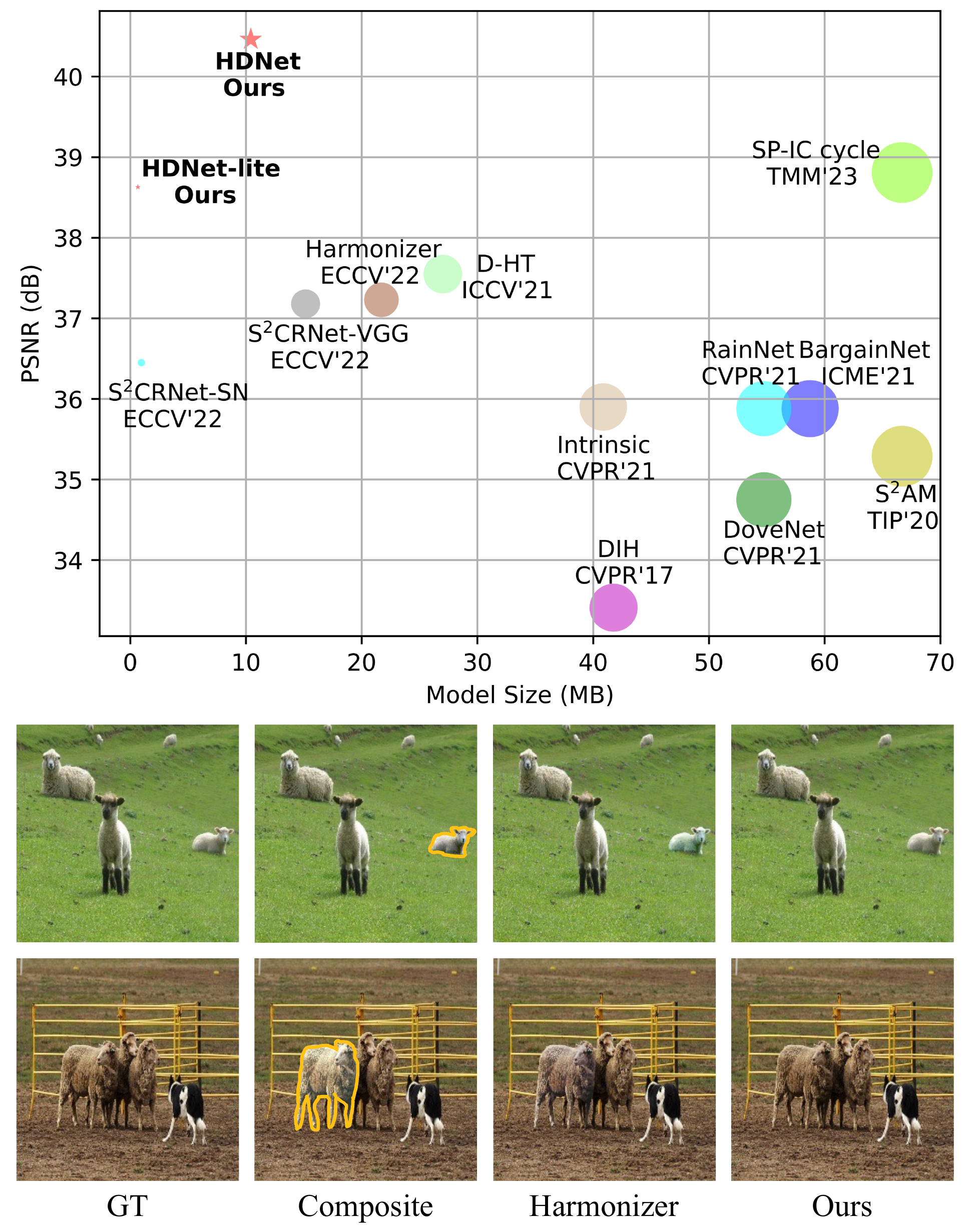}
	\caption{In the top figure, we compare parameter size and performance between our method and other state-of-the-art methods. It can be seen that our method has fewer parameters but achieve state-of-the-art results. In the bottom figure, our method produces a more photorealistic harmonized result.}
	\label{hdih}
\end{figure}

In the computer vision community, integrating image patches from different sources into a cohesive, realistic image is a foundational technique, as demonstrated in image editing~\cite{shi2022spaceedit,ling2021editgan} and scene completion~\cite{cai2021semantic,tang2022not}. However, composite images inevitably encounter inharmonious elements, as foreground and background appearances may have gaps due to varying imaging conditions (e.g., rainy and sunny, morning and dusk). Consequently, image harmonization, which strives for visual consistency within the composite image, constitutes an important and challenging task.

Traditional approaches emphasize the importance of harmonizing hand-crafted low-level appearance statistics, including color statistics~\cite{CT,xue2012TG}, and gradient information~\cite{Mat2016}. Nevertheless, these methods are unsatisfactory in the complex scenarios where the source image exhibits a significant appearance or semantic gap with the target.

With the advances in deep learning, more deep neural network-based methods were proposed~\cite{DoveNet,Bargainnet,Harmonizer,CDTNet}. Most of them use complex network structures or training strategies~\cite{Bargainnet,DoveNet} to accomplish image harmonization tasks. In contrast, color transformation and normalization based models have received extensive attention due to their simplicity and flexibility~\cite{RAIN,CDTNet,liang2021spatial}. 

Color transformation based techniques tend to learn an RGB-to-RGB transformation for image harmonization. For instance, Collaborative Dual Transformation (CDTNet)\cite{CDTNet} incorporates a low-resolution generator for pixel-to-pixel transformation, lookup tables (LUTs) for RGB-to-RGB transformation, and a refinement module that takes advantage of both pixel-to-pixel transformation and RGB-to-RGB transformation.Rencently, Spatial-Separated Curve Rendering Network (S$^2$CRNet)\cite{liang2021spatial} introduces a curve rendering module (CRM) that learns and integrates spatial-specific knowledge, to generate piecewise curve mapping parameters in the foreground region. 

Normalization based methods regard image harmonization as a background-to-foreground style transfer task. Inspired by AdaIN~\cite{AdaIN}, Ling \textit{et al.}~\cite{RAIN} regard image harmonization as a background to foreground style transfer problem and proposed region-aware adaptive instance normalization (RAIN) which captures the style statistics information from the background features and applies it to the foreground. However, as shown in Figure~\ref{hdih}, unexpected patterns still exist and are very severe in some cases.

However, global feature representation and transformation may not be effective enough for image harmonization tasks. This is because different regions of an image have distinct colors, textures, and structural characteristics that cannot be fully captured by global-level features. Another weakness of the above methods is that they use fixed statistics for normalization, which significantly limits their representation ability. Moreover, their model sizes are too large for edge devices, e.g., mobile phones. 

In this paper, we solve the above problems by proposing an efficient dynamic image harmonization network, which hierarchically adapts the features by two dynamics, \emph{i.e.}, local dynamic module and mask-aware global dynamic module from local to global view. To align each foreground local representation with semantically and appearance matched to the background ones, local dynamic module first finds $K$-nearest neighbor background local representations. Then, local dynamic module adaptively reconstruct each foreground local representation by linearly combining it with related background local representations. Besides, for global feature learning, mask-aware global dynamic module utilizes distinct convs for both foreground and background regions. This module enables the model to acquire adaptive representations for these regions and effectively capture their correlations.

As shown in Figure~\ref{hdih}, the proposed framework is efficient and effective compared to existing image harmonization models. Our method achieves higher performance with fewer parameters.

The main contributions can be summarized as follows:
\begin{itemize}
	\item We propose a novel hierarchical dynamic image harmonization network, which adaptively adapt the features from local to the global view for background and foreground visual alignments. 
	
	\item We present a local dynamic module, which finds $K$-nearest neighbor background local representations for each foreground local representation and adjusts the appearance of each foreground local representation.
	
	\item We develop a mask-aware global dynamic module to learn the representations of foreground and background regions as well as their correlations for the global harmonization, leading to better and more efficiently visual consistency.
	
	\item Evaluations on image harmonization datasets demonstrate that our method can achieve state-of-the-art performance using fewer parameters and lower computational costs.
\end{itemize}

\begin{figure*}[t]
	\centering
	\includegraphics[height=12.4cm,width=17.8cm]{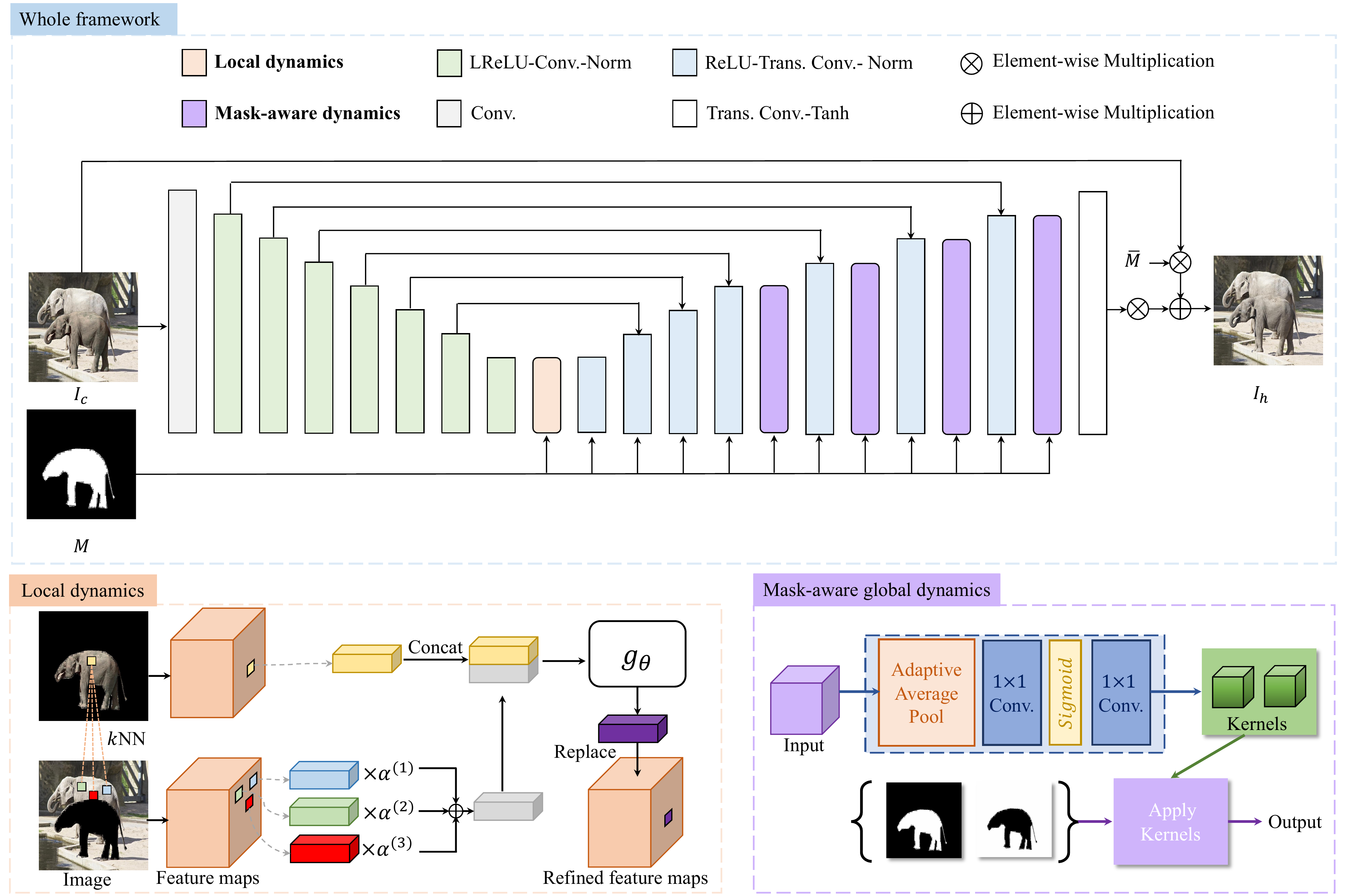}
	\caption{Overview of our proposed hierarchical dynamic image harmonization model. The HDNet consists of an Encoder, a Decoder, a Local Dynamic module and several Mask-aware Global Dynamic modules. Local dynamic module fuses each foreground local representation with its $K$-nearest neighbor background local representations to achieve local visual consistency. Mask-aware global dynamic module aims to learn the representations of foreground and background regions as well as their correlations from the global view, facilitating global visual consistency for the images much more efficiently.}
\end{figure*}

\section{Related Works}
\subsection{Image Harmonization}

Traditional image harmonization methods primarily focus on enhancing composite images by manipulating low-level appearance features. These methods include color transformations to achieve a consistent visual appearance between foreground and background objects~\cite{CT,xue2012TG}, as well as employing multi-scale transformations and statistical analysis to adjust the overall composition~\cite{sunkavalli2010multi}. Although these methods have shown satisfactory results in certain scenarios, they are limited by their reliance on low-level features and may struggle to adapt to more complex situations.

In recent years, deep learning-based approaches have emerged as a powerful alternative for image harmonization, yielding significant improvements in performance~\cite{Bargainnet,DIH,IHT,GuoZJGZ21,CDTNet,Harmonizer,liang2021spatial}. Methods such as DoveNet~\cite{DoveNet} and BargainNet~\cite{Bargainnet} treat image harmonization as a domain translation problem, aiming to enhance domain consistency between the background and foreground objects. By leveraging advanced deep learning techniques, these models can better capture complex relationships and adapt to various scenarios. Ling \textit{et al.}~\cite{RAIN} introduced the Region-aware Adaptive Instance Normalization (RAIN) module, an innovative approach that transfers the statistical properties of background features to normalized foreground features. RAIN has demonstrated promising results, particularly in terms of its ability to harmonize composite images with diverse content. Other methods such as S$^2$CRNet and CDTNet have integrated color transformation into image harmonization tasks, with their models specifically designed to handle high-resolution problems.

Nevertheless, a common limitation of many existing methods is their reliance on global features for transformation, which can be ineffective for image harmonization tasks that require finer control over local details. In response to this challenge, our proposed model adopts a hierarchical approach that progressively adapts features from a local to a global view. By incorporating both local and global information, our model is better suited to harmonize images with varying levels of complexity and detail, ultimately resulting in more natural and visually coherent composites.

\subsection{Style Transfer}
Style transfer is a technique that seeks to modify the stylistic appearance of an image based on given style patterns while maintaining the original content structure. This process has been the focus of numerous studies, leading to the development of various methods that achieve style transfer in different ways.

Huang \textit{et al.}~\cite{AdaIN} proposed Adaptive Instance Normalization (AdaIN), which applies channel-wise mean and variance of style features to align the distribution of content and style images as closely as possible. This method allows for effective style transfer while preserving the underlying content structure. Another approach, Batch-Instance Normalization (Batch-IN)\cite{BatchInsN}, combines the concepts of batch normalization and instance normalization to achieve style transfer. Jing \textit{et al.}\cite{DyIN} introduced dynamic instance normalization, a technique that generates weights using a learnable network that takes the style image as input. This method offers a more flexible and adaptive way to transfer styles. In the Whitening and Coloring Transform (WCT)\cite{wct} method, style transfer is achieved by first whitening the content representation and subsequently coloring it with the style representation. This approach allows for a more direct manipulation of the content and style features. The Style Attentional Network (SANet)\cite{SANet} focuses on efficiently and flexibly integrating local style patterns based on the semantic spatial distribution of the content image. This approach results in a more coherent and natural-looking style transfer. The recent RainNet method~\cite{RAIN} has demonstrated the effectiveness of AdaIN~\cite{AdaIN} in image harmonization tasks. By employing style transfer techniques, RainNet effectively harmonizes the appearance of composite images while preserving their content structure. This approach highlights the potential of style transfer methods in enhancing image harmonization and underscores the need for further research into the interplay between these two fields.

\subsection{Dynamics in Computer Vision}

Dynamic networks~\cite{han2021dynamic,woo2018cbam,drconv,deconv,yaohuimm} focus on improving the representation ability of deep models by adapting model structures or parameters during inferences. At their core, dynamic models adaptively apply different weights to parameters or features of the models based on the input data. This adaptability results in increased model capacity and representation ability, enabling the models to better capture complex relationships and patterns in the data. Attention modules~\cite{woo2018cbam} are typical examples of dynamic networks, where attention maps are calculated to focus on the important channel or salient region. However, adapting important individually for each pixel may lose the translation invariance of the CNN. To solve this problem, Dynamic Region-Aware Convolution~\cite{drconv} is
proposed to assign multiple convolutional filters to different regions separately and share the same filters in each region. Moreover, Deformable Convolution~\cite{deconv,adap_deconv,deconv2} added an offset variable to the position of each sampling point in the convolution kernel, which can realize random sampling nearly points.

\section{Methodology}
Our goal is to learn a hierarchical dynamic network for image harmonization. To achieve this goal, we introduce two sub-modules for improving the performance of basic networks, \emph{i.e.}, local dynamic (LD) module and mask-aware global dynamic (MGD) module.

\subsection{Overview}
Image harmonization aims to adjust the appearance of the foreground object to make it compatible with the background. In a typical image harmonization task, we are given a foreground image $I_f$ and a background image $I_b $. The foreground mask, denoted by $M$, indicates the region to be harmonized in the composite image. The composite image $I_c$ can be formulated as $I_c = M \times I_f + (1 - M) \times I$. It is worth noting that the background mask can be represented as $\overline{M} = 1 - M$.
Following~\cite{sofiiuk2021foreground}, we only employ the foreground MSE loss as our loss function:
\begin{equation}
	\mathcal{L}(I, I_h) = \frac{\sum\limits_{y,x} ||I^{y,x}-{I_h}^{y,x}||}{{\rm Max}\{A_{min}, \sum\limits_{y,x}M^{y,x}\}} .
\end{equation}
$A_{min}$ is a hyperparameter for preventing instability during training and $I^{y,x}$ is the ground truth. 

Figure 2 shows the overall framework of our method. Following~\cite{AdaIN,Bargainnet,CDTNet,Harmonizer}, we employ the U-Net structure as our generator $G$ to harmonize the foreground. The generator $G$ contains an encoder $E$, a decoder $H$, a local dynamic module and several mask-aware global dynamic modules. The details of our network structure can be found in supplementary.
\subsection{Local Dynamic Module}
Considering the significantly variant appearances of different regions of the
foreground and the background, recent methods with global feature alignment may not be effective enough in image harmonization tasks. Thus a background location that exhibits greater similarity to the foreground location require more attention, as they play a crucial role in achieving visually coherent harmonization results. To achieve this, we propose a local dynamic module that can adaptive adjust the appearance of each foreground location by matching and fusing related background locations.

For each image, after passed through the encoder, we can obtain its deep features $F\in \mathbb{R}^{C \times H \times W}$ and the corresponding resized foreground mask $M \in \mathbb{R}^{C \times H \times W}$, where $C$, $H$, $W$ indicate the number of channels, height, and width of $F$, respectively. The encoder feature $F$ can be viewed as a set of $H \times W$ $C$-dimensional local representations. By utilizing the mapping relationship of the mask, these local representations can also be divided into foreground local representations $F_f \in \mathbb{R}^{C \times N_f}$ and background local representations $F_b \in \mathbb{R}^{C \times N_b}$, where $N_f$ is the number of foreground local representations, $N_b$ the number of background local representations and $N_f + N_b = HW$. For each foreground local representations,  we aim to find background local representations with similar appearance and semantics and use these background local representations to adjust its appearance. We first calculate  similarity map $S$ as below:
\begin{gather}
	S^{(i,j)} =  {\rm cos}(F_f^{(i)},F_b^{(j)}),\\
	{\rm cos}(F_f^{(i)},F_b^{(j)}) = \frac{{F_f^{(i)}}^{\top} F_b^{(j)}}{||F_f^{(i)}||\cdot||F_b^{(j)}||},
\end{gather}
where $i \in \{1, ..., N_f\}$, $j \in \{1, ..., N_b\}$, $S^{(i,j)}$ is the distance
between the $i$-th local representation of the foreground image and the $j$-th local representation of background image and $cos(\cdot, \cdot)$ is the cosine similarity. For each foreground local representations, we select its $K$-nearest neighbors in background, and fuse these local representations to one reference representation $\phi_{ref}^{(i)}$:
\begin{equation}
	\phi_{ref}^{(i)} = \sum_{k=1}^{K} \alpha^{(k)} \times F_b^{(k)},
\end{equation}
where $\sum_{k=1}^{K} \alpha^{(k)} = 1$, $\alpha^{(k)} > 0$ and $k=1,...,K$. The $\alpha^{(k)}$ is obtained by applying the softmax function to the selected $K$ local representations. Then we use these reference local representations to adaptively adjust the corresponding foreground local representations. We concatenate the foreground and reference local representations together and fuse them through an adaptive layer:
\begin{equation}
	\phi_{fuse}= g_{\theta}({\rm Concat}(\phi_{ref}, F_f)),
\end{equation}
where $\phi_{fuse}$ is the fused local representations and $g_{\theta}$ indicates the adaptive layer.

\begin{table*}[t]
	\centering
	\resizebox{\textwidth}{!}{
		\begin{tabular}{cccccccccccc}
			\toprule
			\multirow{2}{*}{Model} & \multirow{2}{*}{Param.} & \multicolumn{2}{c}{HAdobe5k} & \multicolumn{2}{c}{HFlickr} & \multicolumn{2}{c}{HCOCO} & \multicolumn{2}{c}{Hday2night}& \multicolumn{2}{c}{Average} \\ \cline{3-12}&&MSE$\downarrow$&PSNR$\uparrow$&MSE$\downarrow$&PSNR$\uparrow$&MSE$\downarrow$&PSNR$\uparrow$&MSE$\downarrow$&PSNR$\uparrow$&MSE$\downarrow$&PSNR$\uparrow$\\
			\midrule
			Composite&-&345.54&28.16&264.35&28.32&69.37&33.94&109.65&34.01&172.47&31.63\\
			DIH~\cite{DIH}&41.76MB&92.65&32.28&163.38&29.55&51.85&34.69&82.34& 34.62&76.77& 33.41\\
			S$^{2}$AM~\cite{SSAM}&66.70MB&63.40&33.77&143.45&30.03&41.07&35.47&76.61&34.50& 59.67&34.35 \\
			iS$^{2}$AM~\cite{sofiiuk2021foreground}&66.70MB& 21.60&38.28&69.43&33.65&16.15 &39.40&40.39&37.87& 24.13&38.41\\
			DoveNet~\cite{DoveNet}&54.76MB& 52.32&34.34&133.14&30.21&36.72& 35.83&51.95&35.27&52.33&34.76\\
			RainNet~\cite{RAIN}&54.75MB &43.35&36.22 &110.59&31.64 &29.52&37.08&57.40&34.83&40.29& 36.12  \\
			BargainNet~\cite{Bargainnet}&58.74MB& 39.94 &35.34&97.32 & 31.34 & 24.84&37.03& 50.98 &35.67& 37.82 &35.88 \\
			Intrinsic~\cite{GuoZJGZ21} &40.86MB&43.02&35.20&105.13&31.34&24.92& 37.16 &55.53& 35.96 &38.71&35.90 \\
			D-HT~\cite{IHT}&27.00MB &38.53&36.88&74.51&33.13&16.89& 38.76&53.01&37.10&30.30&37.55\\
			Harmonizer~\cite{Harmonizer} &21.70MB&21.89&37.64&64.81&33.63&17.34& 38.77&33.14& 37.56&24.26&37.84 \\
			S$^2$CRNet-SN~\cite{liang2021spatial}&\textbf{\textcolor{blue}{0.95MB}} &44.52&35.93 &115.46&31.63&28.25& 37.65 &53.33& 36.28 &43.20&36.45\\
			S$^2$CRNet-VGG~\cite{liang2021spatial}&15.14MB &34.91&36.42 &98.73 &32.48 &23.22& 38.48&51.67& 36.81&35.58&37.18 \\
			SCS-Co~\cite{hang2022scs}&-&21.01 &38.29 &55.83& 34.22 &13.58 & 39.88 &41.75& 37.83 & 21.33 & 38.75\\
			DCCF~\cite{DCCF}&-&23.34 &37.75 &64.77 &33.60&17.07& 38.66&55.76& 37.40&24.65&37.87  \\
			CDTNet~\cite{CDTNet} &-&20.62 &38.24&68.61&33.55 &16.25 & 39.15&36.72& 37.95 & 23.75 & 38.23\\
			SP-IC cycle~\cite{cai2023structure}& 66.70MB & \textbf{\textcolor{blue}{18.17}} & 38.91 & 68.85 &33.88&14.82& 39.73&41.32& 37.90&22.47&38.81  \\
			LEMaRT~\cite{liu2023lemart}& - & 18.80 &\textbf{\textcolor{blue}{39.40}} & \textbf{\textcolor{red}{40.70}} &\textbf{\textcolor{blue}{35.30}} &\textbf{\textcolor{red}{10.10}}& \textbf{\textcolor{blue}{41.00}} & 42.30& 38.10& \textbf{\textcolor{blue}{16.80}} &\textbf{\textcolor{blue}{39.80}}  \\
			\midrule
			HDNet-lite &\textbf{\textcolor{red}{0.65MB}}&24.94 & 39.16 & 63.55& 34.30 & 17.33 &39.21 & \textbf{\textcolor{blue}{32.73}} & \textbf{\textcolor{blue}{38.36}} &24.99& 38.63\\
			HDNet&10.41MB&\textbf{\textcolor{red}{13.58}}&\textbf{\textcolor{red}{41.17}} & \textbf{\textcolor{blue}{47.39}} & \textbf{\textcolor{red}{35.81}} & 
			\textbf{\textcolor{blue}{11.60}} & \textbf{\textcolor{red}{41.04}} & \textbf{\textcolor{red}{31.97}} & \textbf{\textcolor{red}{38.85}} & \textbf{\textcolor{red}{16.55}} & \textbf{\textcolor{red}{40.46}} \\
			\bottomrule
	\end{tabular}}
	\caption{Quantitative comparison across four sub-datasets of iHarmony4~\cite{DoveNet}. Top two performance are shown in \textbf{\textcolor{red}{red}} and \textbf{\textcolor{blue}{blue}}. $\uparrow$ means the higher the better, and $\downarrow$ means the lower the better.}
	\label{tab:my_label}
\end{table*}

\begin{figure*}[t]
	\centering
	\includegraphics[height=16cm,width=17cm]{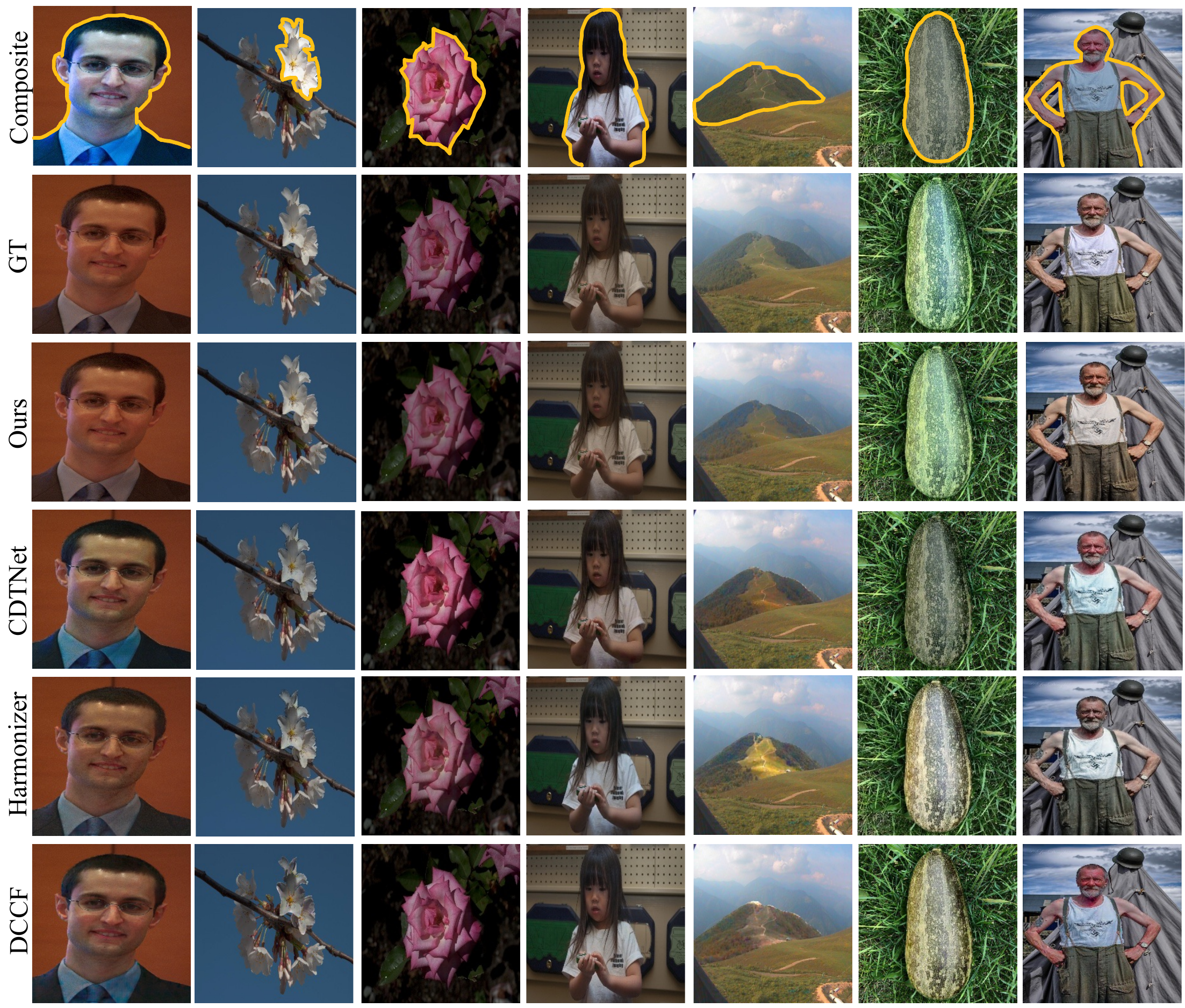}
	\caption{Qualitative comparison on samples from the testing dataset of iHarmony4. The yellow border lines indicate the foreground.}
	\label{fig3}
\end{figure*}

\begin{figure*}[t]
	\centering
	\includegraphics[height=6.8cm,width=16cm]{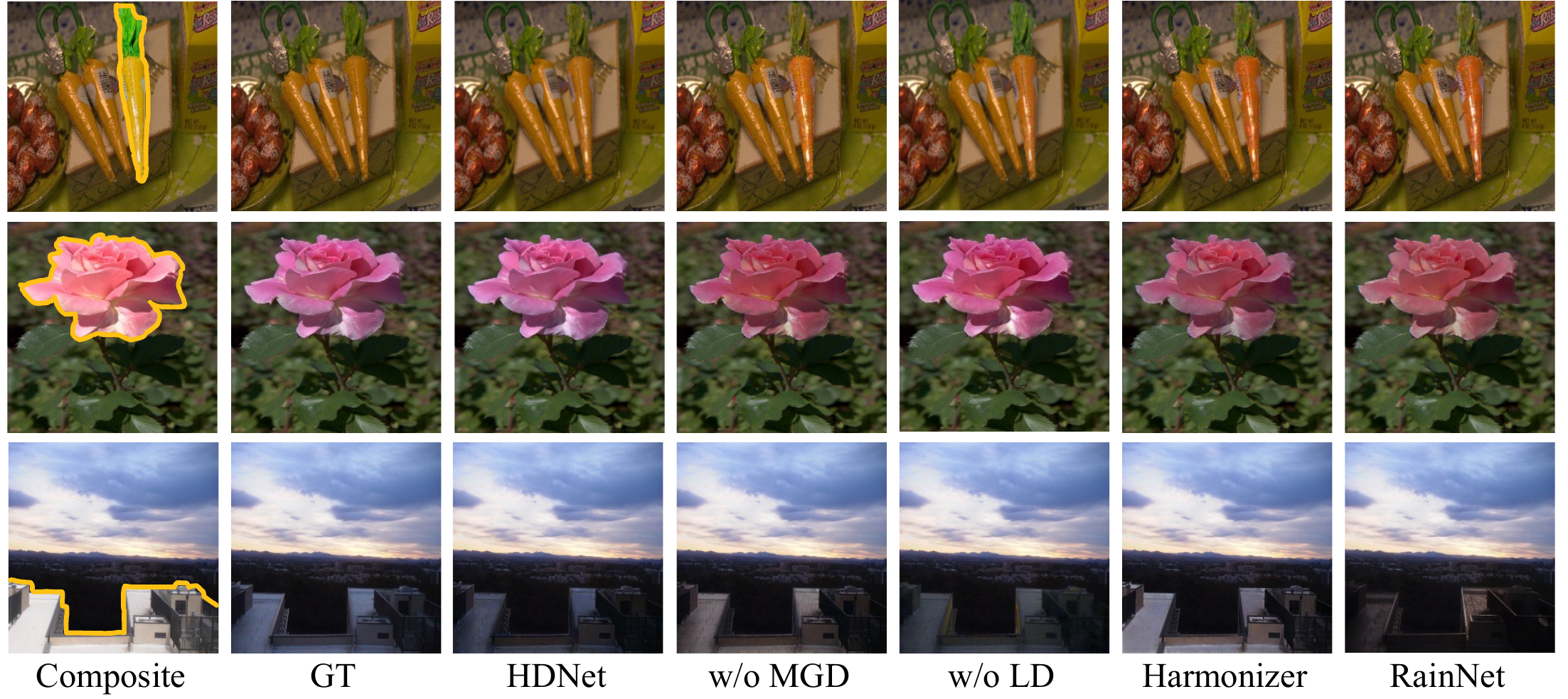}
	\caption{Ablation study on samples from the testing dataset of iHarmony4.}
	\label{fig3}
\end{figure*}

\begin{table*}[t]
	\centering
	\begin{tabular}{cccccccccc}
		\toprule
		\multirow{2}{*}{Model} &\multirow{2}{*}{Param.} & \multicolumn{2}{c}{0\% $\sim$5\%} & \multicolumn{2}{c}{5\% $\sim$15\%} & \multicolumn{2}{c}{15\%$\sim$100\%} &  \multicolumn{2}{c}{Average} \\ \cline{3-10}
		&&MSE$\downarrow$&fMSE$\downarrow$&MSE$\downarrow$&fMSE$\downarrow$&MSE$\downarrow$&fMSE$\downarrow$&MSE$\downarrow$&fMSE$\downarrow$\\
		\midrule
		Composite&-&28.51&1208.86&119.19&1323.23&577.58&1887.05 &172.47&1387.30\\
		S$^{2}$AM~\cite{SSAM}&66.70MB&13.51&509.41&41.79&454.21&137.12&449.81&48.00&481.79 \\
		iS$^{2}$AM~\cite{sofiiuk2021foreground}&66.70MB& 6.35& 288.19&19.69 &226.00& 71.68&235.30&  24.13&260.22\\
		DoveNet~\cite{DoveNet}&54.76MB&14.03&591.88 & 44.90&504.42&152.07&505.82&52.36&549.96\\
		RainNet~\cite{RAIN}&54.75M&11.66&550.38& 32.05 &378.69&117.41&389.80&40.29&469.60\\
		
		BargainNet~\cite{Bargainnet}&58.74MB& 10.55 & 450.33 & 32.13 & 359.49 & 109.23 & 353.84 & 37.82 & 405.23\\ 
		Intrinsic~\cite{GuoZJGZ21} &40.86MB&9.97& 441.02&31.51&363.61 &110.22 & 354.84 &38.71& 400.29 \\
		S$^2$CRNet-SN~\cite{liang2021spatial}&\textbf{\textcolor{blue}{0.95M}} &8.42 &301.97& 29.74& 336.24& 126.56& 405.13& 43.21 &336.99\\
		S$^2$CRNet-VGG~\cite{liang2021spatial}&15.14MB &6.80& \textbf{\textcolor{blue}{239.94}}& 25.37& 271.70 &103.42 &333.96 &35.58 &274.99 \\
		SP-IC cycle~\cite{cai2023structure}& 66.70MB &\textbf{\textcolor{blue}{6.08}} &276.59 & \textbf{\textcolor{blue}{18.27}}&\textbf{\textcolor{blue}{209.56}}& \textbf{\textcolor{blue}{66.44}}& \textbf{\textcolor{blue}{216.37}}&\textbf{\textcolor{blue}{22.47}}&\textbf{\textcolor{blue}{245.75}}  \\
		\midrule
		HDNet-lite&\textbf{\textcolor{red}{0.65MB}}& 6.45&289.71&18.70& 217.66& 76.56&243.58&24.99& 260.65\\
		HDNet&10.41MB&\textbf{\textcolor{red}{4.35}}&\textbf{\textcolor{red}{199.56}}&\textbf{\textcolor{red}{12.82}}&\textbf{\textcolor{red}{150.45}}&\textbf{\textcolor{red}{49.18}}&\textbf{\textcolor{red}{159.82}}&\textbf{\textcolor{red}{16.55}}&\textbf{\textcolor{red}{179.49}}\\
		\bottomrule
	\end{tabular}
	\caption{We measure the error of different methods in foreground ratio range based on the whole test set. fMSE indicates the mean square error of the foreground region.}
	\label{tab:my_label}
\end{table*}

\begin{table*}[t]
	\centering
	\begin{tabular}{ccccccccccc}
		\toprule
		\multirow{2}{*}{Model} & \multicolumn{2}{c}{HAdobe5k} & \multicolumn{2}{c}{HFlickr} & \multicolumn{2}{c}{HCOCO} & \multicolumn{2}{c}{Hday2night}& \multicolumn{2}{c}{Average} \\ \cline{2-3} \cline{4-11}&MSE$\downarrow$&PSNR$\uparrow$&MSE$\downarrow$&PSNR$\uparrow$&MSE$\downarrow$&PSNR$\uparrow$&MSE$\downarrow$&PSNR$\uparrow$&MSE$\downarrow$&PSNR$\uparrow$\\
		\midrule
		Composite&345.54&28.16&264.35&28.32&69.37&33.94&109.65&34.01&172.47&31.63\\
		Base
		& 24.33 &38.83 & 65.77 & 34.05 & 17.51 & 39.27 &34.08 & 38.15&25.20&38.53\\
		+LD& 14.79& \textbf{\textcolor{blue}{40.96}}& 51.51 & 35.57 & 12.44 & 40.82 & 32.79 & 38.45 & 17.64&40.23 \\
		+SA& 15.89 & 40.46 & 55.46 & 35.33 & 13.08 & 40.38 & 32.30 & 39.00 & 19.19&39.81 \\
		+MGD&16.37 &40.81 &55.56 &35.36 &12.82 & 40.75 & 32.44& \textbf{\textcolor{blue}{38.82}}&18.99&40.13 \\ 
		HDNet$^*$& \textbf{\textcolor{blue}{14.39}}&41.03 &\textbf{\textcolor{blue}{48.46}} &\textbf{\textcolor{blue}{35.79}} & \textbf{\textcolor{blue}{11.94}}&\textbf{\textcolor{blue}{40.91}} &\textbf{\textcolor{red}{30.69}}&38.80 & \textbf{\textcolor{blue}{17.08}}&\textbf{\textcolor{blue}{40.34}} \\
		
		\midrule
		HDNet&\textbf{\textcolor{red}{13.58}}&\textbf{\textcolor{red}{41.17}} & \textbf{\textcolor{red}{47.39}} & \textbf{\textcolor{red}{35.81}} & 
		\textbf{\textcolor{red}{11.60}} & \textbf{\textcolor{red}{41.04}} & \textbf{\textcolor{blue}{31.97}}& \textbf{\textcolor{red}{38.85}}& \textbf{\textcolor{red}{16.55}} & \textbf{\textcolor{red}{40.46}} \\
		\bottomrule
	\end{tabular}
	\caption{Ablation study on iHarmony4~\cite{DoveNet}. HDNet$^*$ indicates that we use the learned mask instead of the original mask provided by datasets. SA indicates the spatial attention module used in RainNet~\cite{RAIN} and SCS-Co~\cite{hang2022scs}}
\end{table*}

\begin{table}[t]
	\centering
	\begin{tabular}{cccc}
		\toprule
		Model&Parms.&Time(s)&PSNR$\uparrow$\\\midrule
		S$^{2}$AM~\cite{SSAM}&66.70M&0.25&34.35 \\
		DoveNet~\cite{DoveNet}&54.76M&0.05&34.76\\         BargainNet~\cite{Bargainnet}&58.74M&0.21&35.88 \\ 
		Intrinsic~\cite{GuoZJGZ21} &40.86M&1.17&35.90 \\
		S$^2$CRNet-SN~\cite{liang2021spatial}&\textbf{\textcolor{blue}{0.95M}} &\textbf{\textcolor{red}{0.03}}&\textbf{\textcolor{blue}{36.45}}\\
		\midrule
		HDNet-lite&\textbf{\textcolor{red}{0.65M}}&\textbf{\textcolor{blue}{0.04}}&\textbf{\textcolor{red}{38.63}}\\
		\bottomrule
	\end{tabular}
	\caption{Average processing time on the CPU.}
\end{table}

\subsection{Mask-aware Global Dynamic Module}
Recent works~\cite{RAIN,Bargainnet,SSAM} show that using attention blocks in the decoder helps improve performance. However, it may not be effective to perform spatial attention on hybrid encoder-decoder features since pixel-level adaptation is unsuitable for low-level texture features. 

To learn adaptive representations for harmonious and inharmonious regions, we propose the mask-aware global dynamic module to predict the adaptive convolutional kernels with the guidance of the mask. As shown in Figure 2, mask-aware global dynamic module are incorporated into our networks to better integrate the local information for modeling the visual coherence. Unlike DRconv~\cite{drconv} focusing on the local information, which is unreliable in the harmonious and inharmonious regions, we learn different kernels according to the foreground mask. For efficiency, different groups of filters for foreground and background are applied for the whole input to get the dynamic features. Then the dynamic features are multiplied by the mask. Finally, a summation is applied to obtain the final results for the MGD:
\begin{equation}
	F'_m = (F_m\odot W_f )\otimes M + (F_m\odot W_b )\otimes \overline{M},
\end{equation}
where $\odot$ denotes covolution operation, $W_f$ and $W_b$ are the filters.

\textbf{Why would our model work?} Since LD integrates foreground local representations with the $K$-nearest neighbors of background local representations, effectuating adaptive transfer of the background appearance to the foreground. Concurrently, MGD applies different kernels on the foreground and background regions, each region can be regarded as being assigned an individual decoder to learn the harmonization mapping, but without introducing extra computational cost, since all regions share the same encoder for feature extraction.

\section{Experiments}
In this section, we first introduce the datasets, metrics, and implementation details for our experiments. We then compare HDNet with existing image harmonization methods. We further conduct ablation experiments to evaluate the effectiveness of individual modules in HDNet. Finally, we demonstrate the advantages of HDNet in real-world image harmonization applications.
\subsection{Experiment Setting}

\noindent
\textbf{Datasets.} Following the recent works~\cite{Bargainnet,RAIN,DoveNet}, we conduct image harmonization tasks on iHarmony4 benchmark~\cite{DoveNet}. iHarmony4 includes 73,146 image pairs for image harmonization and contains four subsets: HAdobe5k, HFlickr, HCOCO, and Hday2night. Each sample in iHarmony4 consists of a natural image, a foreground mask, and a composite image (with the foreground generated by GAN~\cite{GAN}). We follow the same partition settings of iHarmony4 as DoveNet~\cite{DoveNet}. Note that we conduct high-resolution (i.e., $1024\times1024$ and $2048\times2048$) experiments on HAdobe5k since only HAdobe5k contains high-resolution images.

\noindent
\textbf{Implementation Details.}
HDNet is trained from scratch by Adam optimizer with $\beta_1=0.9$, and $\beta_2=0.999$. The batch size is set to 12 and we train our HDNet for 120 epochs. The initial learning rate is set to 0.001. The initial learning rate is multiple by 0.1 in the 100-th and 110-th epochs. All images are resized to $256\times256$, batch size set to 12, and no data augmentations are adopted. We use PyTorch~\cite{paszke2019pytorch} to implement our models with Nvidia Tesla A100 GPUs.

\noindent
\textbf{Evaluation.} During the test phase, we use Mean Square Error (MSE), foreground MSE (fMSE), Structural SIMilarity (SSIM), and Peak Signal-to-Noise Ratio (PSNR) to evaluate the performance. To illustrate performance, we qualitatively compare our method with numerous state-of-the-art methods, including DIH~\cite{DIH}, S$^{2}$AM~\cite{SSAM}, iS$^{2}$AM~\cite{sofiiuk2021foreground}, DoveNet~\cite{DoveNet}, RainNet~\cite{RAIN}, Bargainnet~\cite{Bargainnet}, Intrinsic~\cite{GuoZJGZ21}, D-HT~\cite{IHT}, CDTNet~\cite{CDTNet}, Harmonizer~\cite{Harmonizer}, DCCF~\cite{DCCF}, S$^{2}$CRNet~\cite{liang2021spatial}, SCS-CO~\cite{hang2022scs}, SP-IC cycle~\cite{cai2023structure}, INR\cite{inr} and LEMaRT~\cite{liu2023lemart}. 

\subsection{Comparison with Other Methods}

\noindent
\textbf{Performances on different sub-datasets.} Table 1 lists the quantitative results of previous state-of-the-art methods and our method. From Table 1, we can observe that our method outperforms all of them across all sub-datasets and all metrics. From Table 1, we can see that our method achieves best results and also on most sub-data sets. Specifically, compared to the most recent CVPR'23 method LEMaRT~\cite{liu2023lemart} on iHarmony4 dataset, our HDNet brings 0.35 improvement in terms of MSE, and 0.66 dB improvement in terms of PSNR. 

Moreover, to make our model practical, that is, it can be used on edge devices (e.g., mobile phones), we propose HDNet-lite. HDNet-lite is obtained by reducing the number of HDNet channels to $1/4$. Compared to the method with equivalent performance, our HDNet-lite has fewer parameters. For example, compared to RainNet and BarginNet, HDNet-lite only uses 1.2\% of the parameters to achieve better performance in the PSNR metric, demonstrating the effectiveness of the proposed network.

\noindent
\textbf{Influence of foreground ratios.} Following~\cite{RAIN}, we examine the influence of different foreground ratios on the harmonization models, i.e., 0\% to 5\%, 5\% to 15\%, 15\% to 100\%, and overall results. The results of all previous methods and our HDNet are given in Table 2. It can be observed that our HDNet achieves the best performance among all approaches. HDNet works well at various foreground scales, thanks to its combination of hierarchical dynamics.

\noindent
\textbf{Qualitative comparisons.} We take a closer look at model performance and provide qualitative comparisons with the previous competing methods. From the
sample results in Figure 3, it can be easily observed that our approach integrates the foreground objects into
the background image, achieving much better visual consistency than other methods. Our HDNet can achieve these photorealistic results because our HDNet adaptively adjusts the feature of foreground and background by hierarchical dynamics learning.

\subsection{Ablation Study}

\noindent
\textbf{Visual comparison.} To further illustrate the effectiveness of our hierarchical dynamics, we show some output results of ablation experiments in Figure 4. It can be found that compared with the distortion results produced by the baseline, after adding our proposed dynamics, the color and lighting of the output results are close to the real images. Each dynamics contribute to the final result because they conduct dynamic learning at different feature levels. 

\begin{table}[t]
	\centering
	\begin{tabular}{cccccc}
		\toprule
		Model&Resolution&PSNR$\uparrow$&MSE$\downarrow$&fMSE$\downarrow$&SSIM$\uparrow$\\\midrule
		Composite&\multirow{8}{*}{1024$\times$1024}&352.05&28.10&2122.37&0.9642\\
		DoveNet~\cite{DoveNet}&&34.81&51.00& 312.88&0.9729\\
		S$^{2}$AM~\cite{SSAM}&&35.68&47.01&262.39&0.9784\\
		Intrinsic~\cite{GuoZJGZ21}&&34.69&56.34&417.33& 0.9471\\
		RainNet~\cite{RAIN}& &36.61&42.56&305.17& 0.9844\\
		CDTNet~\cite{CDTNet}&&\textbf{\textcolor{blue}{38.77}}&\textbf{\textcolor{blue}{21.24}}& \textbf{\textcolor{blue}{152.13}}&0.9868\\
		INR~\cite{inr}&&38.38&22.68&187.97&\textbf{\textcolor{blue}{0.9886}}\\
		HDNet&&\textbf{\textcolor{red}{41.56}}&\textbf{\textcolor{red}{13.24}}&\textbf{\textcolor{red}{102.53}}&\textbf{\textcolor{red}{0.9931}}\\
		\midrule
		CDTNet~\cite{CDTNet}&\multirow{3}{*}{2048$\times$2048}&37.66&29.02& 198.85&0.9845\\
		INR~\cite{inr}&&\textbf{\textcolor{blue}{38.35}}&\textbf{\textcolor{blue}{24.08}}&\textbf{\textcolor{blue}{192.20}}&\textbf{\textcolor{blue}{0.9886}}\\
		HDNet&&\textbf{\textcolor{red}{41.29}}&\textbf{\textcolor{red}{18.35}}&\textbf{\textcolor{red}{147.25}}&\textbf{\textcolor{red}{0.9911}}\\
		\bottomrule
	\end{tabular}
	\caption{High-resolution experiments on HAdobe5K.}
\end{table}

\noindent
\textbf{Effectiveness of local dynamic module.} Our local dynamic module adjusts the appearance of each foreground local representation according to the $K$-nearest background local representations. In Table 3, we can see that adding LD to the baseline brings 1.7 dB and 7.56 average performance improvement in terms of PSNR and MSE. Moreover, if we remove LD from HDNet, the PSNR will decrease by 0.23 dB and MSE will decrease by 2.44.

\noindent
\textbf{Effectiveness of mask-aware global dynamic module.} Our mask-aware global dynamic module integrates the local information to model visual coherence. Adding MGD to the model will bring significant improvement, proving that it is not effective enough to perform spatial attention~\cite{RAIN,hang2022scs} on hybrid encoder-decoder features since pixel-level adaptation is unsuitable for such low-level texture features. Moreover, if we use the learned mask to replace the original mask, the performance will decline, indicating that the learned mask is unreliable.

\noindent
\textbf{Harmonization performance on CPU.} Our HDNet shows relatively fast processing speed on CPU devices, which enables our method to run on the device side without any cloud computation. To this end, we compare the proposed HDNet with other baseline methods~\cite{SSAM,DoveNet,Bargainnet,IHT,liang2021spatial} in harmonizing under the same experimental environment (Intel Xeon Platinum 8369B CPU on Ubuntu 18.04). The evaluations are conducted on the 50 images in the HAdobe5k sub-dataset and we present the average processing time in Table 4. The experimental results show that our method has the second fastest inference speed when inference on CPU but the performance of our model is much better than other methods.

\noindent
\textbf{High-resolution results.}
Following~\cite{CDTNet}, we conduct high-resolution image harmonization experiments. As shown in Table 5, we can see that our method outperforms all of them across all metrics. Compared with the most recent method INR~\cite{inr}, under $1024\times1024$ resolution setting, our method achieves a huge average performance gain of 9.44 in MSE, 85.44 in fMSE, 0.0045 in SSIM, and 3.18 in PSNR.

\begin{figure}[t]
	\centering
	\includegraphics[height=4.6cm,width=7.6cm]{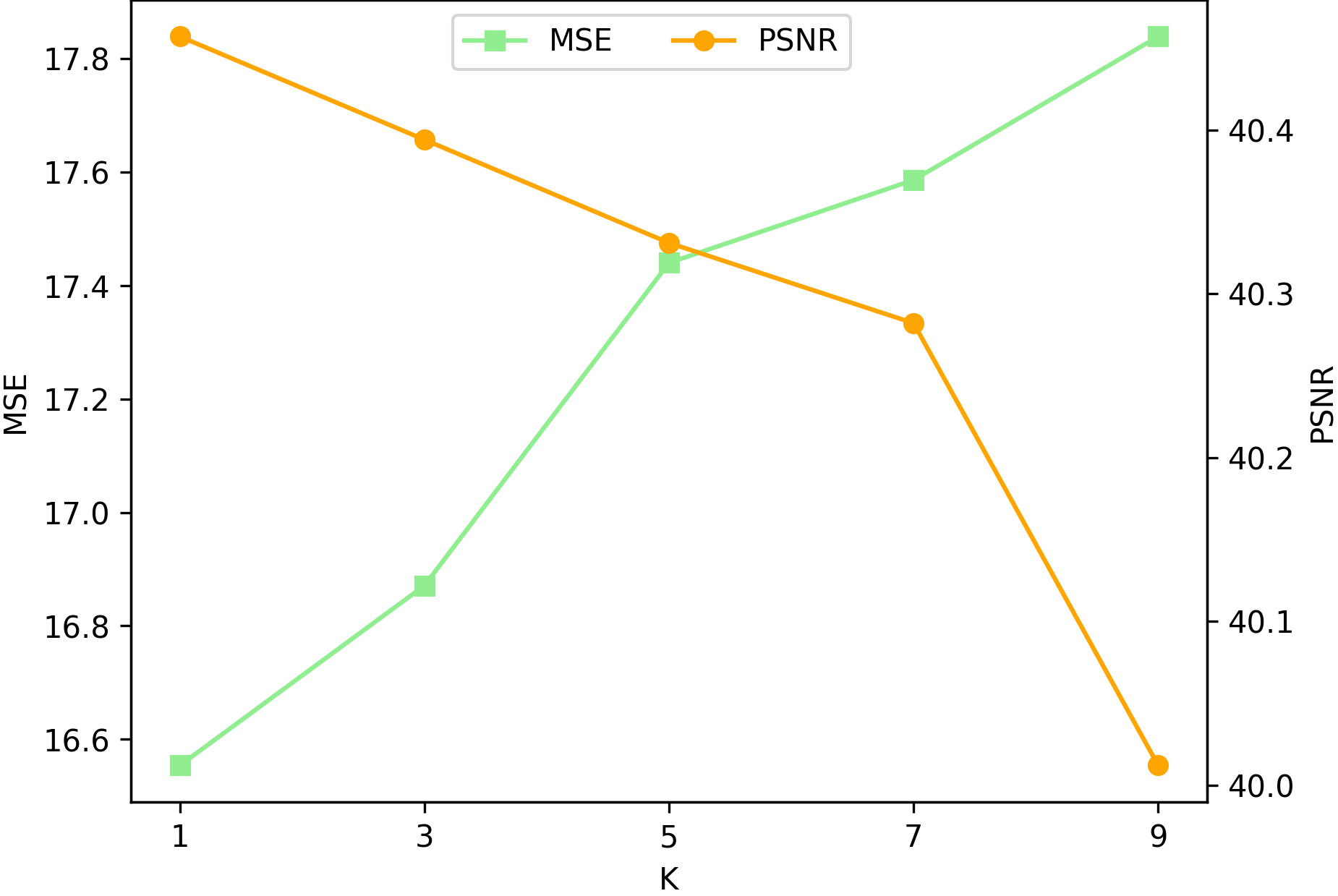}
	\caption{Influence of neighbors. The PSNR score decreases and MSE increases with larger $K$.}
	\label{k}
\end{figure}

\noindent
\textbf{Influence of neighbors.} In the local dynamic module, we need to find the $K$-nearest neighbors in the background image for each local representation of the foreground image. Next, we fuses every foreground local representations with its $K$-nearest background local representations. How to choose a suitable hyperparameter $k$ is thus a key. For this purpose, we perform a low-resolution harmonization task (i.e., $256\times 256$) by varying the value of $K \in \{1, 3, 5, 7, 9\}$, and show the results in Figure~\ref{k}. It can be observed that the performance is best when $K$ is equal to 1. This may be attributed to the provision of potentially negative information by an excessive number of local representations.

\section{Conclusion}
This paper proposes a hierarchical dynamic network (HDNet) from local to global that gradually builds local dynamic module and mask-aware global dynamic module. Local dynamic module fuses each foreground local representation with its $K$-nearest neighbor background local representations to achieve local visual consistency. Mask-aware global dynamic module aims to learn representations of foreground and background regions as well as their correlations from the global view, facilitating global visual consistency for the images much more efficiently. Our method achieves state-of-the-art performances on the benchmark dataset iHarmony4 and our lightweight version model HDNet-lite achieves state-of-the-art results compared to other methods while only has 0.65MB parameters. Our limitation mainly lies in
the dependency on mask. When a disharmonious image does not provide a mask, the performance of using a learning mask is not good. Future investigation into these issues should be required.

\bibliographystyle{ACM-Reference-Format}
\bibliography{sample-base}

\end{document}